\documentclass{article}
\usepackage{spconf,amsmath,graphicx}
\usepackage{booktabs}
\usepackage{xcolor}

\usepackage{caption}
\usepackage{subcaption}

\usepackage{bm}
\usepackage{color, colortbl}
\usepackage{tablefootnote}
\usepackage{tikz}
\usepackage{textpos}

\usepackage[belowskip=0pt,aboveskip=5pt]{caption}

\title{On Compressing Sequences for Self-Supervised Speech Models}
\newcommand*{\affmark}[1][*]{\textsuperscript{#1}}

\name{\parbox{\textwidth}{\centering Yen Meng\affmark[1]$^*$,
Hsuan-Jui Chen\affmark[1]$^*$,
Jiatong Shi\affmark[2],
Shinji Watanabe\affmark[2], \\
Paola Garcia\affmark[3],
Hung-yi Lee\affmark[1],
Hao Tang\affmark[4]
\thanks{$^*$Equal contribution.}}}

\address{
\affmark[1]National Taiwan University,
\affmark[2]Carnegie Mellon University,
\affmark[3]Johns Hopkins University,\\
\affmark[4]The University of Edinburgh}

\begin{document}
\maketitle

\begin{abstract}
Compressing self-supervised models has become increasingly necessary, as self-supervised models become larger.
While previous approaches have primarily focused on compressing the model size,
shortening sequences is also effective in reducing the computational cost.
In this work, we study fixed-length and variable-length subsampling along the time axis in self-supervised learning.
We explore how individual downstream tasks are sensitive to input frame rates.
Subsampling while training self-supervised models not only improves the overall performance on downstream tasks under certain frame rates,
but also brings significant speed-up in inference.
Variable-length subsampling performs particularly well under low frame rates.
In addition, if we have access to phonetic boundaries, we find no degradation in performance for an average frame rate as low as 10 Hz.
\end{abstract}

\begin{keywords}
self-supervised learning, sequence length compression, variable-length subsampling
\end{keywords}

\begin{textblock*}{\textwidth}(0cm, 10.5cm)
\tiny\noindent Copyright 2023 IEEE. Published in the 2022 IEEE Spoken Language Technology Workshop (SLT) (SLT 2022), scheduled for 19-22 January 2023 in Doha, Qatar. Personal use of this material is permitted. However, permission to reprint/republish this material for advertising or promotional purposes or for creating new collective works for resale or redistribution to servers or lists, or to reuse any copyrighted component of this work in other works, must be obtained from the IEEE. Contact: Manager, Copyrights and Permissions / IEEE Service Center / 445 Hoes Lane / P.O. Box 1331 / Piscataway, NJ 08855-1331, USA. Telephone: + Intl. 908-562-3966.
\end{textblock*}

\section{Introduction}
Large-scale self-supervised speech models \cite{w2v2,DeCoAR2.0,hubert,Wavlm,data2vec}, have proven their generality, serving various downstream tasks, such as speech recognition, speaker recognition, and emotion recognition \cite{superb}.
Given the success of large-scale models (Transformers with 12 layers or more) \cite{w2v2,DeCoAR2.0,hubert,Wavlm,data2vec}, it is natural to believe that large model capacity is necessary for the models to be general purpose.
To what extent the large capacity is needed is an open question, but the high computation cost and memory usage pose a high technical barrier for the adoption, and more importantly, training of these models.
In this work, we aim to tackle the high computational cost of self-supervised learning.

To reduce the technical requirement of using these models, several studies have explored model compression techniques, such as weight pruning, knowledge distillation, and architectural modification \cite{lottery,distillation,nas-rl}.
However, in speech processing, the number of frames, along the time axis, is typically the dominating factor at runtime.
For Transformers in particular, though quadratic memory consumption of self-attention is possible to avoid \cite{sa-o(n2)}, most implementations still require quadratic memory and quadratic runtime. 
Sub-quadratic attention mechanisms are actively being developed \cite{povey2018time,longformer,big-bird}, but the overhead typically makes these attention mechanisms less useful in practice \cite{efficient-trans}.
Instead of reducing the runtime and memory complexity of self-attention, we study subsampling techniques to reduce the sequence length directly.

Subsampling (and sometimes concatenating) contiguous frames is a common technique for speeding up training and inference of automatic speech recognizers \cite{Vanhoucke2013,Miao2015,LAS}.
This approach can be as simple as concatenating every two frames \cite{LAS} or dropping every even frames \cite{bahdanau2016end,joint-ctc}.
Subsampling can also be achieved with pooling or convolution with a stride larger than 1 \cite{BH2014, KWL2016, zhang2017very,hori17_interspeech}.
In this paper, we term this family of approaches \textbf{fixed-length subsampling}.
Fixed-length subsampling has been explored in self-supervised learning \cite{SEW,St-SEW,fithubert}.
Both Wu et al.\ \cite{SEW} and Vyas et al.\ \cite{St-SEW} use fixed-length subsampling, paired with upsampling, during the optimization of self-supervised losses. Lee et al.\ \cite{fithubert} is the most similar to our work in that it also uses fixed-length subsampling to reduce the time resolution paired with upsampling in a knowledge distillation setting.
The subsampled representation is then upsampled when evaluated on downstream tasks.
Though this approach achieves the desired computational speed-up, how subsampling affects the learned representation largely remains unclear and is the central focus of this paper. Another difference is that we directly use the subsampled representation in downstream, and explored a framework that works better without upsampling involved.

Since phones in speech come in varying duration, it is widely accepted that self-supervised models enforcing piecewise constant constraints would learn better representations \cite{ACPC,SCPC,dieleman2021variable}.
Piecewise constant constraints can have various forms, but they mostly involve finding boundaries of segments and pooling frame representations into segment representations.
We term this approach \textbf{variable-length subsampling}.
In this work, we study both the computational speed-up and the impact on learned representations with variable-length subsampling.
Our approach involves adding a layer of Continuous Integrate-and-Fire (CIF) \cite{CIF} to learn subsampled representation while optimizing loss functions.

To push the boundary of small, yet competitive self-supervised models, we build our work on top of DistilHuBERT \cite{distilhubert}, a model consisting of two Transformer layers, trained with knowledge distillation from HuBERT \cite{hubert}.
We evaluate our approach on phone recognition, automatic speech recognition (ASR) with word pieces of different sizes, keyword spotting, intent classification, speaker identification, and emotion recognition. 
We find that phone recognition and ASR are the tasks most impacted by the subsampled representation.
Matching the frame rate of downstream tasks gives the best performance.
Variable-length subsampling has particularly strong performance for the settings with a low frame rate.
For analysis, we also show that given the phonetic segmentation, variable-length sampling can be as good as, if not better than, the DistilHuBERT baseline while having a frame rate as low as 10 Hz.

\section{Compressing Sequences with Subsampling}

The model we study in this paper is based on DistilHuBERT.
For an input waveform $s$, there is a convolutional neural network (CNN) that converts the waveform $s$ into a sequence of feature vectors $x_1, \dots, x_T$, or simply $x_{1:T} = \text{CNN}(s)$.
we obtain a sequence of hidden vectors $u_1, \dots, u_T$ from the teacher model $f_\text{H}$ (in this case HuBERT).
Similarly, $v_{1:T} = f_\text{D}(x_{1:T})$, where $f_\text{D}$ is the student model, DistilHuBERT.
The student model is trained to minimize the distance of the representations $L(u_{1:T}, v_{1:T}; W) = \sum_{t=1}^T d(u_t, W v_t)$, where $W$ is a trainable projection and the distance $d$ can be cosine, dot product, Euclidean distance, or a combination of them.
In DistilHuBERT, the distillation is applied to multiple layers of HuBERT via multitask learning.
In particular, we minimize the sum of several loss terms $L(u^{(\ell)}_{1:T}, v_{1:T}; W^{(\ell)})$, where $u^{(\ell)}_{1:T}$ is the $\ell$-th hidden layer of HuBERT and each loss has a learnable projection $W^{(\ell)}$ (also called a prediction head).

\begin{figure*}
    \centering
    \includegraphics[width=0.95\linewidth]{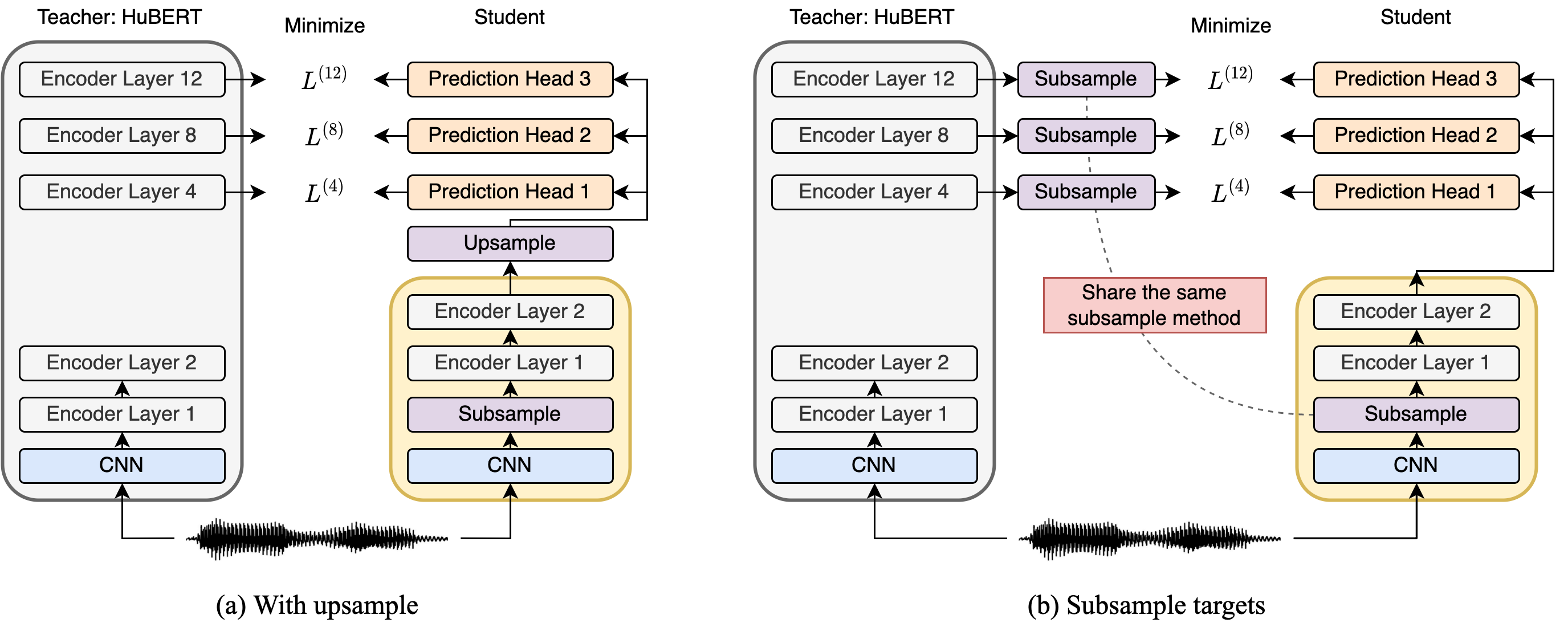}
    \caption{Two subsampling options. (a) A subsampling module is placed after the student's CNN layer, and an upsampling module is placed after the student's encoder output. (b) Identical subsampling modules are placed after the student's CNN layer and after the teacher's encoder output.}
    \label{fig:subsampling}
\end{figure*}

To subsample frames in time while maintaining the use of the loss from DistilHuBERT, we have two options, one with subsampling followed by upsampling, and another by subsampling the targets.
Specifically, the first option minimizes
\begin{align}
    L\Big(u_{1:T}, \text{upsample}(f_{\text{D}}(\text{subsample}(x_{1:T})))\Big)
\end{align}
by having upsampling after the output of the student, while the second option optimizes
\begin{align}
    L\Big(\text{subsample}(u_{1:T}), f_{\text{D}}(\text{subsample}(x_{1:T}))\Big)
\end{align}
by subsampling the output of the teacher.
Figure \ref{fig:subsampling} shows the two options.
For the rest of the sections, we will discuss the choices of subsampling and upsampling functions.

\subsection{Fixed-Length Subsampling}
\label{sec:fixed-length}

For fixed-length subsampling, we adopt two commonly used approaches, convolution subsampling and average pooling  with a stride larger than 1.
We can control the frame rates by changing the strides. For convolution subsampling, the kernel size is set to the same as the stride size.

\subsection{Variable-Length Subsampling}
\label{sec:variable-length}

For variable-length subsampling, we follow a two-step approach. A segmentation (a sequence of boundaries) is first proposed; vectors within each segment are pooled (for example with weighted averaging) and passed to the subsequent layers.
We use Continuous Integrate-and-Fire (CIF) \cite{CIF} to produce segmentation proposals.
At a high level, for an input sequence of length $T$, the CIF module takes input from the previous layer and produces a sequence $\alpha_1, \alpha_2, ..., \alpha_T$ of nonnegative numbers (using a combination of convolution, feedforward layer, and sigmoid function).
Whenever $\sum_{i=1}^t \alpha_i$, i.e., the accumulation up to time $t$ crosses an integer boundary, a segment boundary at time $t$ is proposed (or fired).
In other words, the sequence $\alpha_{1:T}$ controls both where and how many boundaries should be present.
In fact, fixed-length subsampling can be seen as a special case with $\alpha_t = 1/F$ for all $t=1, \dots, T$ where $F$ can be set based on the desired subsampling rate.

The CIF module can be trained end to end without any supervision. In practice, however, the segmentation seldom corresponds well with any actual boundaries in speech. Below we discuss several options of boundary guidance to help CIF learn more meaningful boundaries.

\vspace{0.5em}
\noindent\textbf{Cardinality Guidance}\hspace{0.25em} The cardinality guidance encourages the CIF module to produce the desired number of segments at training time; hence the name.
Specifically, we add the term
\begin{align}
L_\text{card}(\alpha) = \left( \frac{\sum_{i=1}^T \alpha_i - K}{T} \right)^2.
\end{align}
to our loss function, where $K$ is the desired number of segments.
The normalization $T$ (absent in the original CIF formulation) is added to make sure the term is comparable across utterances of different lengths.
Though the guidance is used at training time, at test time, we use the $\alpha$'s produced by the CIF module as is.

\vspace{0.5em}
\noindent\textbf{Segmentation Guidance}\hspace{0.25em} In some cases where we have access to phonetic boundaries, such as through forced alignments or other unsupervised approaches, the segmentation can be used as a source of supervision for the CIF module.

Suppose a segmentation of $K$ segments is provided as a sequence of boundaries in time indices $t_1, \dots, t_K$.
To encourage the boundaries to be placed in accordance with the provided target, we introduce a segment-based loss
\begin{align}
L_\text{seg}(\alpha) &= \sum_{k=1}^K \left| \sum_{j=1}^{t_k} \alpha_j - k \right|.
\end{align}
The loss only focuses on where the boundaries are proposed.
We also introduce a more stringent constraint, constructing a target $\alpha_i^\text{sup} = \frac{1}{t_{k+1} - t_k}$ for $t_k \leq i < t_{k+1}$.
We optimize
\begin{align}
L_\text{frame}(\alpha, \alpha^{\text{sup}}) &= \| \alpha - \alpha^{\text{sup}} \|_1
\end{align}
to make sure the sequence $\alpha$ produced by the CIF module is close to the target $\alpha^\text{sup}$ at every frame; hence a frame-based loss.

Each of these losses can be added to our loss function with an interpolation factor.
The segmentations for supervision are only used at training time, while at test time, we use the $\alpha$'s produced by the CIF module as is.

\section{Experimental Setting}
Our experiment is based on the original DistilHuBERT implementation with S3PRL \cite{superb} and fairseq \cite{fairseq}. The pre-training setting and hyperparameters, including learning rate schedule, are the same as the original DistilHuBERT implementation except that we introduce the subsampling and upsampling modules. We have three prediction heads, targeting the representations of the $4^\text{th}$, $8^\text{th}$, and $12^\text{th}$ layers of a frozen HuBERT model. We use the 960-hour LibriSpeech dataset \cite{librispeeech} for pre-training, and all models are trained for 200,000 updates with a batch size of 24.

The CIF module for variable-length subsampling consists of a single one-dimensional 512-channel convolution with a stride of 1 and kernel width of 5, followed by a feedforward layer of 512 dimensions and an output dimension of 1.

The experiment is evaluated on a subset of the SUPERB benchmark\footnote{https://superbbenchmark.org}, including phone recognition (PR), automatic speech recognition (ASR), keyword spotting (KS), intent classification (IC), speaker identification (SID), and emotion recognition (ER). We further evaluate on an additional task, automatic speech recognition with word pieces (ASR-5k).
The ASR-5k model is trained with 5000 sentencepiece \cite{sentencepiece} targets (trained with byte-pair-encoding \cite{bpe}). Both ASR and ASR-5k are evaluated without a language model (LM). To compare the performance with the DistilHuBERT paper, only the representation of the last layer (without the prediction heads) is used for downstream evaluation.

\section{Preliminary Analysis}

Before we train our models with subsampling, we explore a range of frame rates for several tasks.
We concatenate contiguous hidden vectors produced by the vanilla DistilHuBERT to achieve a target frame rate.
This approach preserves the information sent to the downstream tasks while only changing the frame rate.
We also explore averaging contiguous hidden vectors as an alternative. 
Since KS, IC, SID, and ER are utterance-level tasks, subsampling the hidden vectors does not affect the performance, so we only focus on PR, ASR, and ASR-5k.

Results are shown in Table~\ref{tab:asr-frame-rate}, where two subsampling methods are tested: concatenating (cat) and average pooling (avg).
Phone recognition starts to fail for frame rates lower than 12.5 Hz. ASR with characters completely fails for frame rates lower than 25 Hz, while ASR with word pieces can sustain subsampling to a frame rate as low as 6.25 Hz.

\begin{figure}[t]
    \centering
    \includegraphics[width=0.97\linewidth]{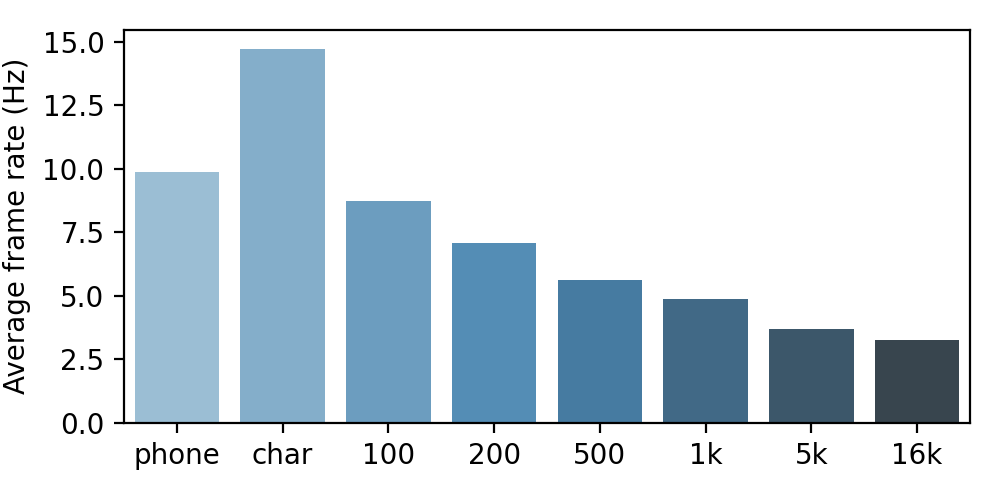}
    \caption{Average frame rates for different units. Phones and characters are labeled as \textit{phone} and \textit{char}. The numbers 100, 200, 500, 1k, 5k, and 16k represent the number of word pieces learned with BPE. The average frame rates are computed on the LibriSpeech \textit{dev-clean} subset.}
    \label{fig:unit-frame-rate}
\end{figure}

Figure \ref{fig:unit-frame-rate} shows the average frame rates of different units, such as phones, characters, and word pieces.
When we increase the number of word pieces learned, the size of the word pieces increases.
In particular, when the number of word pieces is large, many of the word pieces are actual words, and the average frame rate decreases accordingly.
The results in Table~\ref{tab:asr-frame-rate} are consistent with the average frame rates in Figure~\ref{fig:unit-frame-rate}, as these tasks rely on a CTC layer that can only produce labels as many as it has frames for.

We also find that averaging is on par with concatenation, and decide to use averaging as the pooling method for the rest of the experiments.

\colorlet{40ms}{blue!10}
\colorlet{70ms}{teal!10}
\colorlet{80ms}{teal!10}
\colorlet{90ms}{yellow!10}
\colorlet{160ms}{orange!10}

\begin{table}[t]
    \small
    \centering
    \caption{Results of subsampling the output of DistilHuBERT. FP and FR represent the frame period and frame rate, respectively, after subsampling. The metrics include phone error rate (PER) and word error rate (WER).}
    \label{tab:asr-frame-rate}
    \begin{tabular}{l|cc|ccc}
    \toprule
    \textbf{Model} & \textbf{FP} & \textbf{FR} & \textbf{PR} & \textbf{ASR} & \textbf{ASR-5k}\\
     & ms & Hz & PER $\downarrow$ & WER $\downarrow$ & WER $\downarrow$\\
     \midrule
    DistilHuBERT & 20 & 50 & 16.27 & 13.37 & 12.86 \\
     \midrule
    \rowcolor{40ms} cat 2 & 40 & 25 & 15.17 & 17.44 & 13.01 \\
    \rowcolor{80ms} cat 4 & 80 & 12.5 & 31.03 & $>100$ & 13.77 \\
    \rowcolor{160ms} cat 8 & 160 & 6.25 & 82.32 & $>100$ & 17.83\\
    \midrule
    \rowcolor{40ms} avg 2 & 40 & 25 & 15.57 & 16.45 & 12.95 \\
    \rowcolor{80ms} avg 4 & 80 & 12.5 & 30.12 & $>100$ & 14.06 \\
    \rowcolor{160ms} avg 8 & 160 & 6.25 & 79.57 & $>100$ & 18.86\\
    \bottomrule
    \end{tabular}
\end{table}

\section{Experiments}

\begin{figure*}[t]
    \centering
    \begin{subfigure}[b]{0.33\linewidth}
        \centering
        \includegraphics[width=\linewidth]{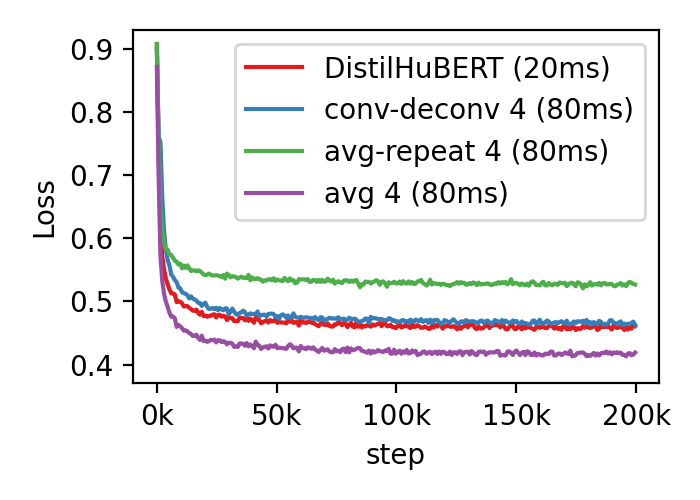}
    \end{subfigure}
    \hfill
    \begin{subfigure}[b]{0.33\linewidth}
        \centering
        \includegraphics[width=\linewidth]{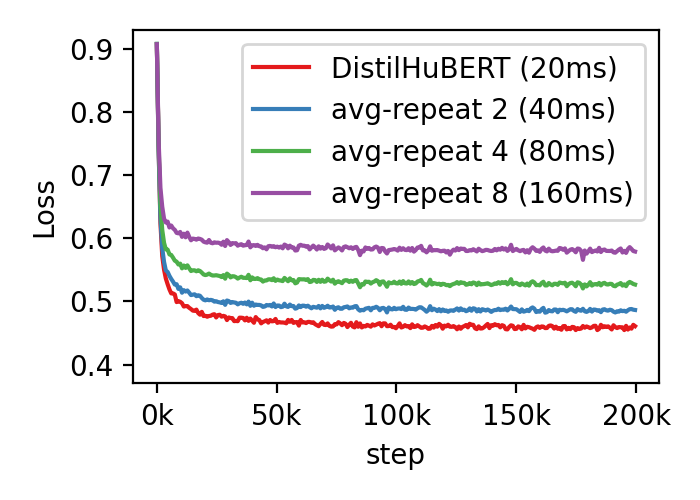}
    \end{subfigure}
    \hfill
    \begin{subfigure}[b]{0.33\linewidth}
        \centering
        \includegraphics[width=\linewidth]{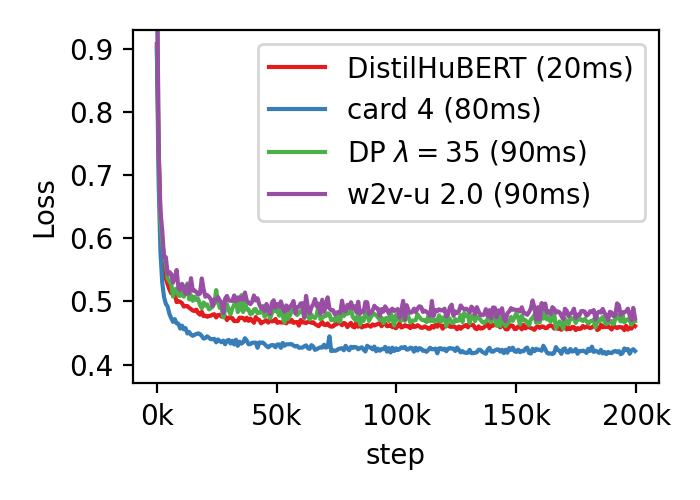}
    \end{subfigure}
    
    \caption{Losses for pretraining with subsampling. \textit{Left}: subsampling and upsampling pairs for pretraining compared to DistilHuBERT and subsampling both the teacher and the student. \textit{Middle}: various frame rates of using averaging and repeating for subsampling and upsampling. \textit{Right}: various types of boundary guidance.}
    \label{fig:pretraining-loss}
\end{figure*}

Unlike the previous section, here, we study models pre-trained along with subsampling. Models trained with subsampling could potentially learn to represent the input speech differently from the vanilla DistilHuBERT.

\subsection{Fixed-Length Subsampling}

In this section, we explore the two options for fixed-length subsampling: subsampling paired with upsampling shown in Figure~\ref{fig:subsampling}~(a), and subsampling of both the teacher's and the student's output shown in Figure~\ref{fig:subsampling}~(b).
For subsampling and upsampling pairs, we have convolution paired with deconvolution and averaging paired with repeating (duplicating frames to the desired frame rate).

In Figure~\ref{fig:pretraining-loss} (Left), we show the pretraining losses for different subsampling and upsampling pairs, and in Table~\ref{tab:downstream} (I), we have their respective downstream performance.
In Figure~\ref{fig:pretraining-loss} (Left), we first note that convolution paired with deconvolution is able to recover the DistilHuBERT loss while averaging paired with repeating is worse.
In Figure~\ref{fig:pretraining-loss} (Middle), we also find that more aggressive sampling makes it more difficult to match the training loss of the vanilla DistilHuBERT.
However, in Table~\ref{tab:downstream} (I), the downstream performance for averaging paired with repeating is on par with convolution paired with deconvolution for the two frame rates we explored.

Due to this finding and the simplicity, we use averaging to conduct the experiments for subsampling both the teacher and the student.
The training loss is shown in Figure~\ref{fig:pretraining-loss} (Left) and the downstream performance is in Table~\ref{tab:downstream} (II).
We do find the training loss to be lower (perhaps due to having fewer frames), and the downstream performance is generally on par, if not better than, with upsampling.
These results are also comparable to the ones in Table~\ref{tab:asr-frame-rate},
including the failed case for phone recognition at the frame rate of 6.25 Hz.

\colorlet{40ms}{blue!10}
\colorlet{70ms}{teal!10}
\colorlet{80ms}{teal!10}
\colorlet{90ms}{yellow!10}
\colorlet{160ms}{orange!10}

\newcolumntype{?}{!{\vrule width 1pt}}

\begin{table*}[t]
    \small
    \centering
    \caption{Downstream performance for various subsampling approaches. FP and FR denote the frame period and frame rate, respectively. The metrics include phone error rate (PER), word error rate (WER), and accuracy (Acc). For further results, please refer to the SUPERB website.}
    \label{tab:downstream}
    \begin{tabular}{l|ccccc|cccccc}
    \toprule
        \textbf{Model} & \textbf{FP} & \textbf{FR} & \textbf{Params} & \textbf{MACs} & \textbf{MACs-C} & \textbf{PR} & \textbf{ASR-5k} & \textbf{KS} & \textbf{IC} & \textbf{SID} & \textbf{ER} \\
         & ms & Hz & Millions & GMACs & GMACs & PER $\downarrow$ & WER $\downarrow$ & Acc $\uparrow$ & Acc $\uparrow$ & Acc $\uparrow$ & Acc $\uparrow$ \\
        \midrule
        DistilHuBERT & 20 & 50 & 23.49 & 758.9 & 207.1 & 16.27 & 12.86 & 95.98 & 94.99 & 73.54 & 63.02 \\
        \midrule
        \multicolumn{9}{l}{\textbf{(I) Fixed-Length} 
        \textit{- with upsampling}} \\
        \midrule
        \rowcolor{40ms} conv-deconv 2 & 40 & 25 & 25.20 & 667.6 & 115.8 & 16.06 & 13.62 & 95.78 & 92.06 & 67.13 & 63.08 \\
        \rowcolor{80ms} conv-deconv 4 & 80 & 12.5 & 26.90 & 610.3 & 58.5 & 32.04 & 16.15 & 95.75 & 90.67 & 62.65 & 61.62 \\
        \midrule
        \rowcolor{40ms} avg-repeat 2 & 40 & 25 & 23.49 & 664.6 & 112.8 & 15.54 & 13.50 & 96.04 & 95.15 & 71.16 & 62.98 \\
        \rowcolor{80ms} avg-repeat 4 & 80 & 6.25 & 23.49 & 607.3 & 55.5 & 30.69 & 16.14 & 95.78 & 93.33 & 68.56 & 61.13 \\
        \midrule
        \multicolumn{9}{l}{\textbf{(II) Fixed-Length} 
        \textit{- subsampling targets}} \\
        \midrule
        \rowcolor{40ms} avg 2 & 40 & 25 & 23.49 & 664.6 & 112.8 & 15.43 & 13.31 & 95.59 & 94.31 & 72.19 & 63.17 \\
        \rowcolor{80ms} avg 4 & 80 & 12.5 & 23.49 & 607.3 & 55.5 & 30.50 & 15.55 & 95.88 & 93.12 & 69.33 & 62.41 \\
        \rowcolor{160ms} avg 8 & 160 & 6.25 & 23.49 & 579.5 & 27.7 & 79.98 & 24.90 & 95.26 & 90.85 & 68.46 & 60.79 \\
        \midrule
        \multicolumn{9}{l}{\textbf{(III) Variable-Length} \textit{- subsampling targets}} \\
        \midrule
        \rowcolor{40ms} card 2 & 40 & 25 & 24.81 & 675.7 & 123.9 & 17.23 & 14.24 & 94.97 & 88.74 & 70.44 & 62.04 \\
        \rowcolor{80ms} card 4 & 80 & 12.5 & 24.81 & 620.7 & 68.9 & 38.51 & 16.88 & 95.13 & 90.51 & 70.27 & 61.31 \\
        \midrule
        \rowcolor{40ms} DP $\lambda=0$  & 40 & 25 & 24.81 & 680.6 & 128.8 & 15.53 & 14.19 & 95.65 & 94.54 & 71.42 & 62.67 \\
        \rowcolor{80ms} DP $\lambda=25$ & 80 & 12.5 & 24.81 & 623.3 & 71.5 & 21.94 & 15.33 & 95.72 & 94.62 & 71.05 & 63.19 \\
        \rowcolor{90ms} DP $\lambda=35$ & 90 & 11.1 & 24.81 & 614.6 & 62.8 & 31.73 & 16.66 & 95.72 & 94.41 & 69.46 & 62.38 \\
        \rowcolor{160ms} DP $\lambda=75$ & 160 & 6.25 & 24.81 & 593.4 & 41.6 & 78.41 & 27.04 & 95.39 & 89.16 & 66.83 & 61.64 \\
        \midrule
        \rowcolor{90ms} w2v-u 2.0 & 90 & 11.1 & 24.81 & 616.6 & 64.8 & 23.37 & 15.71 & 95.62 & 94.65 & 71.70 & 61.74 \\
        \midrule
        \multicolumn{9}{l}{\textbf{(IV) Variable-Length} 
        \textit{- subsampling targets}} \\
        \midrule
        MFA & 100 & 10 & - & - & - & 12.33 & 11.85 & - & - & - & - \\
        \bottomrule
        \end{tabular}
        \label{tab:2}
\end{table*}

\subsection{Variable-Length Subsampling}

Based on the findings in the previous section, we decide to perform variable-length subsampling on both the teacher and the student.
Recall that variable-length subsampling is achieved by producing a sequence $\alpha$ that decides where a boundary should be placed.
Here, we choose a weighted sum of the vectors within a segment to produce a segment representation.
In particular, suppose $\alpha_t, \dots, \alpha_{t'}$ are the $\alpha$'s in between two boundaries. The weights used for weighted averaging are $\epsilon, \alpha_t, \dots, \alpha_{t'} - \epsilon'$, where $\epsilon$ is the leftover weight that exceeds the integer from the previous boundary and $\epsilon'$ is the leftover weight that exceeds the integer boundary and gets carried over to the next boundary.
In order to match the subsampling exactly for both the teacher and the student, we decide to reuse the $\alpha$ produced by the student on the teacher, as indicated in Figure~\ref{fig:subsampling} (b).

We first explore cardinality guidance, denote as \textit{card}, where we set the number of segments to match the desired frame rate. For example, the number of segments $K$ is set to $T/4$ if the desired frame rate is 12.5 Hz. We tune the weight of the added loss term $L_\text{card}$ and find that a weight of $0.5$ works the best.
The downstream performance is shown in the first two rows of Table~\ref{tab:downstream} (III).
The results are generally worse than fixed-length subsampling.
We suspect for the cardinality guidance is too weak for constraining the model, so we turn to the stronger segmentation guidance.

Below we explore two options, using smoothed HuBERT codes and unsupervised ASR, to obtain segmentation for guiding the CIF module. 
\subsubsection{Smoothed HuBERT Codes as Guidance}
\label{sec:dp-hubert}
The amount of repeated HuBERT codes tend to correlate well with the duration of phones \cite{lakhotia2021generative}, so we can make use of the codes produced by the teacher as a proxy to segmentation.
Here we run k-means on the output of HuBERT (layer 6, following \cite{lakhotia2021generative}) to obtain the codes.
We then use the dynamic programming algorithm proposed by Kamper and van Niekerk \cite{dp} to obtain the segmentation.
Their algorithm generally smooths the frequent changes of a HuBERT code sequence, while also providing a hyperparameter $\lambda$ penalizing short segments.
In other words, larger $\lambda$ produces larger segments, leading to a lower frame rate.
There is no clear correspondence between $\lambda$ and frame rate (except that they are positively correlated), so we simply sweep $\lambda$ to find the desired frame rates.
Recall that there are a frame-based approach and a segment-based approach to incorporate segmentation guidance.
We tune the weight of the two additional loss terms, and find that $0.005$ and $0.25$ work best for $L_\text{seg}$ and $L_\text{frame}$, respectively. Despite the weight of $L_\text{seg}$ being deceptively small, it is crucial to prevent CIF from producing degenerate solutions.

Results of this approach are listed in the second block of Table~\ref{tab:downstream} (III).
Note that we tune $\lambda$ to find a frame rate closest to the desirable one. For readability, we round the frame rates to the desirable ones in Table~\ref{tab:downstream} (III).
This approach performs significantly better than others for the 12.5 Hz frame rate,
while matching the result of others for the 25 Hz frame rate.
It shows the great potential of variable-length subsampling and calls for better segmentation.

\subsubsection{Unsupervised ASR as Guidance}

Another approach to obtain a segmentation of speech is through unsupervised ASR \cite{baevski2021unsupervised, liu2022towards}.
Our pre-trained unsupervised ASR model is a simplified version of wav2vec-U 2.0 \cite{liu2022towards} that is trained without the auxiliary k-means cluster loss.
The model is trained on audio from the 100-hour LibriSpeech and texts from the LibriSpeech language modeling text.
Once the model is trained, it simply acts as a frame classifier, producing posterior probabilities of phones at a frame rate of 16.7 Hz (60 ms frame period).
The segmentation can be obtained by finding the maximum of each frame as predictions, and merging the same contiguous predictions.
Since voice activity detection is involved during training, we overwrite segments where the silence is detected.
Once the boundaries are extracted, we add frame-based and segment-based losses with the same weights as in Section~\ref{sec:dp-hubert}.

The downstream performance of using the segmentation of wav2vec-U 2.0 is in the last row of Table~\ref{tab:downstream} (III).
We also conduct a similar experiment with the smoothed HuBERT codes for comparison, matching the frame rate of the segmentation produced by the wav2vec-U 2.0 model.
We find that this approach is particularly strong at phone recognition and ASR with word pieces, giving the best performance and frame rate tradeoff.

The training loss of various variable-length subsampling is shown in Figure~\ref{fig:pretraining-loss} (Right).
Though the downstream performance of using wav2vec-U 2.0 segmentation provides a better tradeoff, the loss is slightly higher than the approach of smoothing HuBERT codes.

\subsection{Topline: Subsampling with Forced Alignments}

Given the results from previous sections, an accurate segmentation could prove to be sufficient for an aggressive subsampling frame rate.
As an analysis, we study forced alignments as a variable-length subsampling approach of its own.
Forced alignments (of phones) are obtained from the Montreal Forced Aligner (MFA) \cite{mfa}.
Once the segmentation are obtained, we simply use average pooling to obtain a segment representation from the teacher and the student.
We follow  DistilHuBERT training without adding any other terms.
For the downstream tasks, we also use forced alignments for subsampling.
Note that the CIF module is not used in this setting, as segmentation is always provided.

The result is shown in Table~\ref{tab:2}~(IV) and serves as our topline result.
We do not report downstream performance for all tasks as transcripts are not available for computing forced alignments on those datasets.
This approach achieves an average frame rate of 10 Hz (i.e, the average frame rate of phones), and has the best phone recognition and ASR results.
The result suggests that there is still room for better segmentations.

\section{Discussion}

To measure the impact of subsampling on runtime, we report average multiply-accumulate operations (MACs) on a subset of LibriSpeech \textit{test-clean} consisting of utterances between 1 and 20 seconds. MACs are measured in inference mode without counting prediction heads and upsampling.
MACs of various subsampling approaches are reported in Table~\ref{tab:downstream}.
The improvement in MACs is obscured by the MACs of the seven CNN layers, as the MACs of CNN dominate everything else.
We therefore also report MACs without the CNN layers (denoted as MACs-C).
The improvement is then clear and consistent with the reduction in frame rates.

In summary, we study effect of subsampling in self-supervised models, exploring different input frame rates for various downstream tasks.
ASR with larger word pieces is more resistant to more aggressive subsampling, while ASR with characters can only sustain subsampling up to approximately 25 Hz.
Simple averaging performs the best among all fixed-length subsampling options.
Variable-length subsampling performs slightly behind fixed-length subsampling in high frame-rate settings but performs much better in low frame-rate settings.
Segmentation can be used to guide variable-length subsampling, and heavily impacts the learned representations.

For future work, the effect of subsampling on large self-supervised models, particularly, by training them with subsampling from random initialization, is largely unexplored.
Aggressive subsampling in both supervised and self-supervised settings are also still open.
In fact, it is even possible to encode an utterance into a single vector \cite{HGVL2019}, independent of the input length.
Much research is needed in this direction.

\section{Acknowledgments}
We thank the Taiwan Web Service and the National Center for High-performance Computing (NCHC) of National Applied Research Laboratories (NARLabs) in Taiwan for providing the computing and storage resources.
Part of the work presented here was carried out during the 2022 Jelinek Memorial Summer Workshop on Speech and Language Technologies at Johns Hopkins University, which was supported with unrestricted gifts from Amazon, Microsoft, and Google.

\bibliographystyle{IEEEbib}
\bibliography{strings,refs}

\begin{thebibliography}{10}

\bibitem{w2v2}
Alexei Baevski, Yuhao Zhou, Abdelrahman Mohamed, and Michael Auli,
\newblock ``{wav2vec} 2.0: A framework for self-supervised learning of speech
  representations,''
\newblock in {\em Advances in Neural Information Processing Systems}, 2020.

\bibitem{DeCoAR2.0}
Shaoshi Ling and Yuzong Liu,
\newblock ``{DeCoAR} 2.0: Deep contextualized acoustic representations with
  vector quantization,''
\newblock {\em arXiv preprint arXiv:2012.06659}, 2020.

\bibitem{hubert}
Wei-Ning Hsu, Benjamin Bolte, Yao-Hung~Hubert Tsai, Kushal Lakhotia, Ruslan
  Salakhutdinov, and Abdelrahman Mohamed,
\newblock ``{HuBERT}: Self-supervised speech representation learning by masked
  prediction of hidden units,''
\newblock {\em IEEE/ACM Transactions on Audio, Speech, and Language
  Processing}, vol. 29, pp. 3451--3460, 2021.

\bibitem{Wavlm}
Sanyuan Chen, Chengyi Wang, Zhengyang Chen, Yu~Wu, Shujie Liu, Zhuo Chen, Jinyu
  Li, Naoyuki Kanda, Takuya Yoshioka, Xiong Xiao, et~al.,
\newblock ``{WavLM}: Large-scale self-supervised pre-training for full stack
  speech processing,''
\newblock {\em IEEE Journal of Selected Topics in Signal Processing}, 2022.

\bibitem{data2vec}
Alexei Baevski, Wei-Ning Hsu, Qiantong Xu, Arun Babu, Jiatao Gu, and Michael
  Auli,
\newblock ``data2vec: A general framework for self-supervised learning in
  speech, vision and language,''
\newblock in {\em ICML}, 2022.

\bibitem{superb}
Shu-wen Yang, Po-Han Chi, Yung-Sung Chuang, Cheng-I~Jeff Lai, Kushal Lakhotia,
  Yist~Y. Lin, Andy~T. Liu, Jiatong Shi, Xuankai Chang, Guan-Ting Lin,
  Tzu-Hsien Huang, Wei-Cheng Tseng, Ko-tik Lee, Da-Rong Liu, Zili Huang, Shuyan
  Dong, Shang-Wen Li, Shinji Watanabe, Abdelrahman Mohamed, and Hung-yi Lee,
\newblock ``{SUPERB: Speech Processing Universal PERformance Benchmark},''
\newblock in {\em Interspeech}, 2021.

\bibitem{lottery}
Jonathan Frankle and Michael Carbin,
\newblock ``The lottery ticket hypothesis: Finding sparse, trainable neural
  networks,''
\newblock in {\em ICLR}, 2019.

\bibitem{distillation}
Geoffrey Hinton, Oriol Vinyals, and Jeffrey Dean,
\newblock ``Distilling the knowledge in a neural network,''
\newblock in {\em NIPS Deep Learning and Representation Learning Workshop},
  2015.

\bibitem{nas-rl}
Barret Zoph and Quoc~V. Le,
\newblock ``Neural architecture search with reinforcement learning,''
\newblock in {\em ICLR}, 2017.

\bibitem{sa-o(n2)}
Markus~N. Rabe and Charles Staats,
\newblock ``Self-attention does not need $\mathrm{O}(n^2) $ memory,''
\newblock {\em arXiv preprint arXiv:2112.05682}, 2021.

\bibitem{povey2018time}
Daniel Povey, Hossein Hadian, Pegah Ghahremani, Ke~Li, and Sanjeev Khudanpur,
\newblock ``A time-restricted self-attention layer for {ASR},''
\newblock in {\em ICASSP}. IEEE, 2018, pp. 5874--5878.

\bibitem{longformer}
Iz~Beltagy, Matthew~E. Peters, and Arman Cohan,
\newblock ``Longformer: The long-document transformer,''
\newblock {\em arXiv preprint arXiv:2004.05150}, 2020.

\bibitem{big-bird}
Manzil Zaheer, Guru Guruganesh, Kumar~Avinava Dubey, Joshua Ainslie, Chris
  Alberti, Santiago Ontanon, Philip Pham, Anirudh Ravula, Qifan Wang, Li~Yang,
  et~al.,
\newblock ``Big bird: Transformers for longer sequences,''
\newblock {\em Advances in Neural Information Processing Systems}, vol. 33,
  2020.

\bibitem{efficient-trans}
Yi~Tay, Mostafa Dehghani, Dara Bahri, and Donald Metzler,
\newblock ``Efficient transformers: A survey,''
\newblock {\em CSUR}, 2022.

\bibitem{Vanhoucke2013}
Vincent Vanhoucke, Matthieu Devin, and Georg Heigold,
\newblock ``Multiframe deep neural networks for acoustic modeling,''
\newblock in {\em ICASSP}, 2013.

\bibitem{Miao2015}
Yajie Miao, Mohammad Gowayyed, and Florian Metze,
\newblock ``{EESEN}: End-to-end speech recognition using deep {RNN} models and
  {WFST}-based decoding,''
\newblock in {\em ASRU}, 2015.

\bibitem{LAS}
William Chan, Navdeep Jaitly, Quoc Le, and Oriol Vinyals,
\newblock ``Listen, attend and spell: A neural network for large vocabulary
  conversational speech recognition,''
\newblock in {\em ICASSP}, 2016.

\bibitem{bahdanau2016end}
Dzmitry Bahdanau, Jan Chorowski, Dmitriy Serdyuk, Philemon Brakel, and Yoshua
  Bengio,
\newblock ``End-to-end attention-based large vocabulary speech recognition,''
\newblock in {\em ICASSP}. IEEE, 2016, pp. 4945--4949.

\bibitem{joint-ctc}
Suyoun Kim, Takaaki Hori, and Shinji Watanabe,
\newblock ``Joint {CTC}-attention based end-to-end speech recognition using
  multi-task learning,''
\newblock in {\em ICASSP}. IEEE, 2017, pp. 4835--4839.

\bibitem{BH2014}
Samy Bengio and Georg Heigold,
\newblock ``Word embeddings for speech recognition,''
\newblock in {\em Interspeech}, 2014.

\bibitem{KWL2016}
Herman Kamper, Weiran Wang, and Karen Livescu,
\newblock ``Deep convolutional acoustic word embeddings using word-pair side
  information,''
\newblock in {\em ICASSP}, 2016.

\bibitem{zhang2017very}
Yu~Zhang, William Chan, and Navdeep Jaitly,
\newblock ``Very deep convolutional networks for end-to-end speech
  recognition,''
\newblock in {\em ICASSP}. IEEE, 2017, pp. 4845--4849.

\bibitem{hori17_interspeech}
Takaaki Hori, Shinji Watanabe, Yu~Zhang, and William Chan,
\newblock ``{Advances in Joint {CTC}-Attention Based End-to-End Speech
  Recognition with a Deep {CNN} Encoder and {RNN}-{LM}},''
\newblock in {\em Interspeech}, 2017, pp. 949--953.

\bibitem{SEW}
Felix Wu, Kwangyoun Kim, Jing Pan, Kyu~J. Han, Kilian~Q. Weinberger, and Yoav
  Artzi,
\newblock ``Performance-efficiency trade-offs in unsupervised pre-training for
  speech recognition,''
\newblock in {\em ICASSP}, 2022.

\bibitem{St-SEW}
Apoorv Vyas, Wei-Ning Hsu, Michael Auli, and Alexei Baevski,
\newblock ``{On-demand compute reduction with stochastic wav2vec 2.0},''
\newblock in {\em Interspeech}, 2022, pp. 3048--3052.

\bibitem{fithubert}
Yeonghyeon Lee, Kangwook Jang, Jahyun Goo, Youngmoon Jung, and Hoi~Rin Kim,
\newblock ``{FitHuBERT: Going Thinner and Deeper for Knowledge Distillation of
  Speech Self-Supervised Models},''
\newblock in {\em Interspeech}, 2022, pp. 3588--3592.

\bibitem{ACPC}
Jan Chorowski, Grzegorz Ciesielski, Jaroslaw Dzikowski, Adrian Lancucki, Ricard
  Marxer, Mateusz Opala, Piotr Pusz, Pawel Rychlikowski, and Michal
  Stypulkowski,
\newblock ``Aligned contrastive predictive coding,''
\newblock in {\em Interspeech}, 2021.

\bibitem{SCPC}
Saurabhchand Bhati, Jes{\'u}s Villalba, Piotr Żelasko, Laureano
  Moro-Vel{\'a}zquez, and Najim Dehak,
\newblock ``Segmental contrastive predictive coding for unsupervised word
  segmentation,''
\newblock in {\em Interspeech}, 2021.

\bibitem{dieleman2021variable}
Sander Dieleman, Charlie Nash, Jesse Engel, and Karen Simonyan,
\newblock ``Variable-rate discrete representation learning,''
\newblock {\em arXiv preprint arXiv:2103.06089}, 2021.

\bibitem{CIF}
Linhao Dong and Bo~Xu,
\newblock ``{CIF}: Continuous integrate-and-fire for end-to-end speech
  recognition,''
\newblock in {\em ICASSP}, 2020.

\bibitem{distilhubert}
Heng-Jui Chang, Shu-wen Yang, and Hung-yi Lee,
\newblock ``{DistilHuBERT}: Speech representation learning by layer-wise
  distillation of hidden-unit {BERT},''
\newblock in {\em ICASSP}, 2022.

\bibitem{fairseq}
Myle Ott, Sergey Edunov, Alexei Baevski, Angela Fan, Sam Gross, Nathan Ng,
  David Grangier, and Michael Auli,
\newblock ``{fairseq}: A fast, extensible toolkit for sequence modeling,''
\newblock in {\em NAACL-HLT}, 2019.

\bibitem{librispeeech}
Vassil Panayotov, Guoguo Chen, Daniel Povey, and Sanjeev Khudanpur,
\newblock ``{Librispeech}: An {ASR} corpus based on public domain audio
  books,''
\newblock in {\em ICASSP}, 2015.

\bibitem{sentencepiece}
Taku Kudo and John Richardson,
\newblock ``{SentencePiece}: A simple and language independent subword
  tokenizer and detokenizer for neural text processing,''
\newblock in {\em EMNLP}, 2018.

\bibitem{bpe}
Rico Sennrich, Barry Haddow, and Alexandra Birch,
\newblock ``Neural machine translation of rare words with subword units,''
\newblock in {\em ACL}, 2016.

\bibitem{lakhotia2021generative}
Kushal Lakhotia, Eugene Kharitonov, Wei-Ning Hsu, Yossi Adi, Adam Polyak,
  Benjamin Bolte, Tu-Anh Nguyen, Jade Copet, Alexei Baevski, Abdelrahman
  Mohamed, et~al.,
\newblock ``On generative spoken language modeling from raw audio,''
\newblock {\em Transactions of the Association for Computational Linguistics},
  vol. 9, pp. 1336--1354, 2021.

\bibitem{dp}
Herman Kamper and Benjamin van Niekerk,
\newblock ``Towards unsupervised phone and word segmentation using
  self-supervised vector-quantized neural networks,''
\newblock in {\em Interspeech}, 2021.

\bibitem{baevski2021unsupervised}
Alexei Baevski, Wei-Ning Hsu, Alexis Conneau, and Michael Auli,
\newblock ``Unsupervised speech recognition,''
\newblock {\em Advances in Neural Information Processing Systems}, vol. 34, pp.
  27826--27839, 2021.

\bibitem{liu2022towards}
Alexander~H. Liu, Wei-Ning Hsu, Michael Auli, and Alexei Baevski,
\newblock ``Towards end-to-end unsupervised speech recognition,''
\newblock {\em arXiv preprint arXiv:2204.02492}, 2022.

\bibitem{mfa}
Michael McAuliffe, Michaela Socolof, Sarah Mihuc, Michael Wagner, and Morgan
  Sonderegger,
\newblock ``{Montreal} forced aligner: Trainable text-speech alignment using
  {Kaldi},''
\newblock in {\em Interspeech}, 2017.

\bibitem{HGVL2019}
Albert Haque, Michele Guo, Prateek Verma, and Li~Fei-Fei,
\newblock ``Audio-linguistic embeddings for spoken sentences,''
\newblock in {\em ICASSP}, 2019.

\end{thebibliography}

\end{document}